\documentclass{svproc}
\usepackage{url}

\usepackage{graphicx}
\usepackage{bm}
\usepackage{amsmath,amssymb}
\usepackage{multirow,booktabs}

\begin{document}
\mainmatter              
\title{MTLM: Incorporating Bidirectional Text Information to Enhance Language Model Training in Speech Recognition Systems}
\titlerunning{MTLM}  
%

\author{Qingliang Meng\inst{1} \and 
Pengju Ren\inst{2} \and
Tian Li\inst{4} \and
Changsong Dai\inst{3} \and
Huizhi Liang\inst{4}}
\authorrunning{Q. Meng et al.} 
%
%
\institute{Princeton University, Princeton NJ 08544, USA,\\
\email{I.Ekeland@princeton.edu},\\ WWW home page:
\texttt{http://users/\homedir iekeland/web/welcome.html}
\and
Universit\'{e} de Paris-Sud,
Laboratoire d'Analyse Num\'{e}rique, B\^{a}timent 425,\\
F-91405 Orsay Cedex, France}

\institute{Megatronix, Beijing, China \and
Baidu, Inc., Beijing, China \and
Shumei AI Research Institute, Beijing China \and
Newcastle University, Newcastle Upon Tyne, United Kingdom\\
\email{qingliang.meng@megatronix.co}, \email{renpengju@baidu.com}, \email{t.li56@newcastle.ac.uk}, \email{daichangsong@ishumei.com}, \email{huizhi.liang@newcastle.ac.uk}}

\maketitle              

\begin{abstract}
Automatic speech recognition (ASR) systems normally consist of an acoustic model (AM) and a language model (LM). The acoustic model estimates the probability distribution of text given the input speech, while the language model calibrates this distribution toward a specific knowledge domain to produce the final transcription. Traditional ASR-specific LMs are typically trained in a unidirectional (left-to-right) manner to align with autoregressive decoding. However, this restricts the model from leveraging the right-side context during training, limiting its representational capacity. In this work, we propose MTLM, a novel training paradigm that unifies unidirectional and bidirectional manners through 3 training objectives: ULM, BMLM, and UMLM. This approach enhances the LM's ability to capture richer linguistic patterns from both left and right contexts while preserving compatibility with standard ASR autoregressive decoding methods. As a result, the MTLM model not only enhances the ASR system's performance but also support multiple decoding strategies, including shallow fusion, unidirectional/bidirectional n-best rescoring. Experiments on the LibriSpeech dataset show that MTLM consistently outperforms unidirectional training across multiple decoding strategies, highlighting its effectiveness and flexibility in ASR applications.

\end{abstract}

\section{Introduction}
Automatic Speech Recognition (ASR) is the technique of transcribing spoken language into text \cite{ref3}. In ASR systems, language models (LMs) play a crucial role in improving transcription accuracy by resolving ambiguities that arise from acoustic similarities under different linguistic and contextual backgrounds \cite{ref49,ref50,ref12,ref48,ref51_interspeech2022}. One widely adopted decoding strategy is shallow fusion, which combines the output probabilities of the acoustic model (AM) and LM during left-to-right beam search decoding. To match this autoregressive decoding process, most LMs used in ASR are trained with unidirectional objectives \cite{ref20}.

However, from a language modeling perspective, both left and right contexts are informative for understanding a word or phrase \cite{ref9,ref39,ref49}. Bidirectional training - such as masked language modeling (MLM) - has been shown to be more efficient and effective for general language understanding \cite{ref15}. Despite this, MLM-based LMs are not directly compatible with standard ASR decoding workflows and are typically limited to post-processing calibration scenarios such as n-best rescoring methods \cite{ref20}. Moreover, combining unidirectional and bidirectional training objectives within a single model often results in information conflicts, as their underlying conditional probability assumptions differ significantly \cite{ref5,ref9,ref12,ref39}.

To address these challenges, we propose MTLM, a novel multi-task training paradigm for LM in ASR systems. 
MTLM incorporates three complementary training objectives: (1) Unidirectional LM (ULM) to simulate left-to-right decoding; (2) Bidirectional Masked LM (BMLM) to enhance language modeling with full contextual information; and (3) Unidirectional Masked LM (UMLM), an auxiliary task designed to bridge the informational gap between ULM and BMLM. Meanwhile, it can boost the robustness of autoregressive decoding when there are errors in preceding predictions. This design encourages smoother information sharing among tasks and mitigates training conflicts. Our contributions are threefold:
\begin{itemize}
	\item We introduce bidirectional context into ASR-specific LM training without sacrificing decoding compatibility. The auxiliary UMLM task is key to resolve objective conflicts and enhance predicting robustness, enabling the model to benefit from both directions of context.
	\item The MTLM framework unifies shallow fusion, unidirectional rescoring, and bidirectional rescoring decoding capabilities within a single model, achieving superior performance over traditional training approaches in these 3 decoding scenarios. ASR system with MTLM language model dose not require to have separate task-specific language models, significantly reducing the resource for LM preparation.
	\item We conduct a detailed error analysis across different utterance lengths. MTLM outperforms unidirectional LM in medium and long utterances, with only minor degradation on short utterances, demonstrating its contextual modeling advantages.
\end{itemize}

\section{Related Work}
Several approaches have been proposed to enhance ASR performance using language models. For instance, \cite{ref39} utilize BERT \cite{ref7} and other bidirectional LMs for n-best rescoring. To reduce the computational cost of bidirectional LMs, \cite{ref38} estimates pseudo-log-likelihoods via regression. In a similar vein, \cite{ref9} improves rescoring by employing the ELECTRA model, which offers faster inference compared to BERT. However, these bidirectional models are trained solely with bidirectional objectives and thus cannot be used for shallow fusion decoding. To address this, \cite{ref12} applies knowledge distillation to transfer the conditional probabilities from bidirectional LMs to unidirectional LMs, enabling shallow fusion. Nevertheless, since the conditioning contexts of bidirectional and unidirectional LMs are inherently different, such distillation may introduce inconsistencies and is theoretically unsound.

Additionally, \cite{ref5} train GPT \cite{ref16}, BERT, and other language models independently and perform rescoring by linearly combining their scores. However, this approach requires multiple separately trained LMs with different training objectives to be loaded during decoding, which significantly increases computational cost. In contrast, our model utilizes a unified parameter set capable of both unidirectional and bidirectional language modeling, requiring only a single forward pass to compute LM scores. Moreover, while \cite{ref5} rely on simple score summation for fusion, this explicit and shallow combination may limit the model’s expressiveness. Our method, by contrast, performs implicit fusion during training through carefully designed auxiliary tasks, which bridge the gap between unidirectional and bidirectional objectives. This allows our model to achieve greater flexibility and improved performance in ASR systems.

\section{Methodology}

\begin{figure*}[ht]
\setlength{\abovecaptionskip}{0.1cm}
  \centering
  \includegraphics[width=1.0\textwidth]{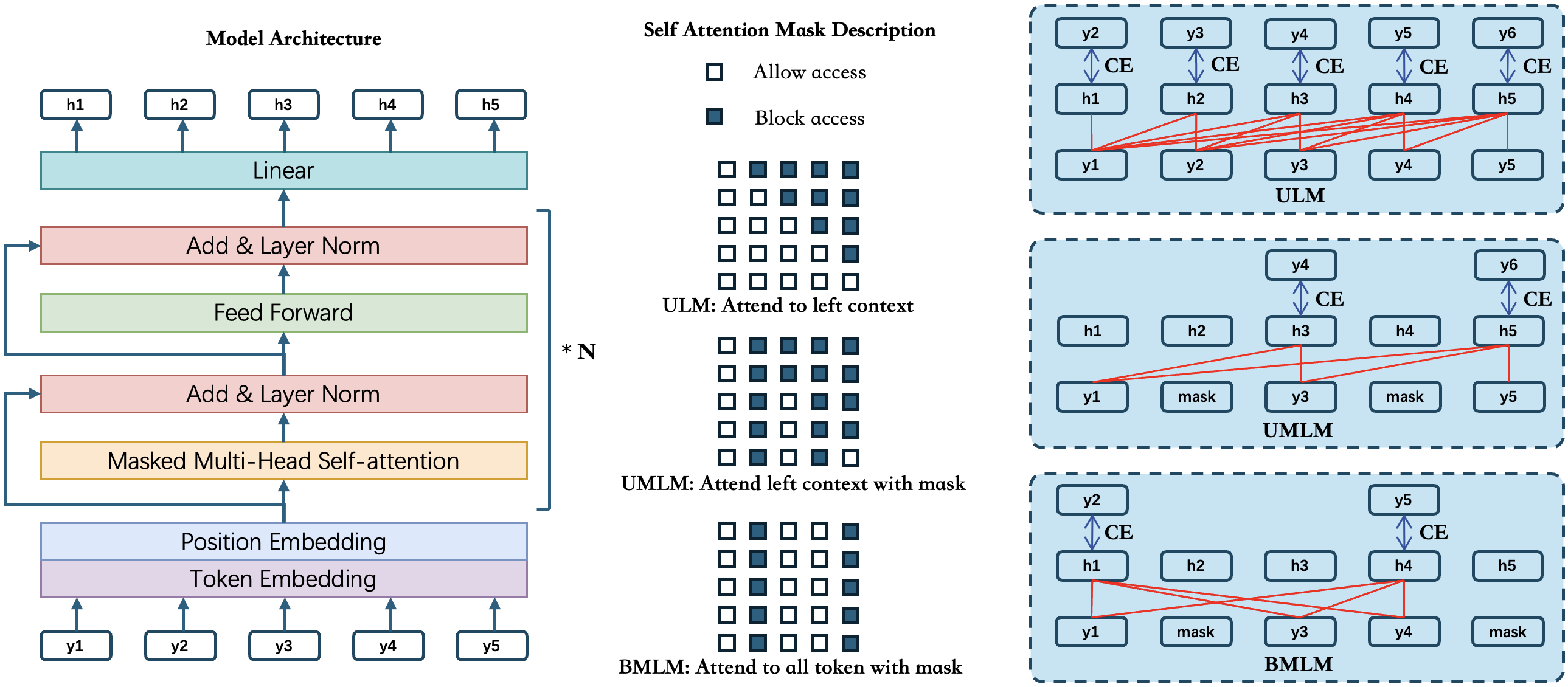}
  \caption{Illustration of MTLM. The left part describes the model structure. The middle part illustrates the mask strategies used by 3 sub-tasks: ULM, UMLM, and BMLM. The right part shows the loss calculations of each sub-tasks.}
  \label{fig:MTLM_model}
\end{figure*}

As shown on the left side of Fig~\ref{fig:MTLM_model}, our model adopts a Transformer-based encoder architecture. The input $\boldsymbol{Y}=y_1,y_2,y_3,...,y_n$ is a sequence of tokens, which is passed through $N$ layers of Transformer blocks to produce a token-level probability distribution $\boldsymbol{H}=h_1,h_2,h_3,...,h_n$. Each Transformer block shares the same architecture, consisting of a masked multi-head attention layer, a feedforward layer, and two layer normalization modules.

Our training framework MTLM incorporates three complementary sub-tasks: Unidirectional Language Modeling (ULM), Bidirectional Masked Language Modeling (BMLM), and Unidirectional Masked Language Modeling (UMLM). Each sub-task has its own masking strategy applied to the attention layer to control the flow of contextual information within the model. These tasks are jointly optimized through a multi-task learning paradigm \cite{ref8,ref43}, where the final training loss is the sum of all sub-task losses.

\subsection{Unidirectional Language Modeling (ULM)}
\label{sec:ULM}
In ASR, the LM is designed to calibrate the score of generated hypothesis, typically expressed as the joint probability $P(\boldsymbol{y})$ of the output token sequence \cite{ref10}. In our framework, the ULM task adopts a left-to-right modeling approach, where the model predicts each token based on its preceding context. The optimization goal is Eq~\ref{eq5}. Here $n$ denotes the length of the token sequence, and $\theta$ represents MTLM model's parameters. 

\begin{equation}
\label{eq5}
\mathcal{L}_{\mathrm{ULM}}\left(\boldsymbol{Y},\boldsymbol{\theta} \right)=\mathbb{E}\left(\sum_{i=1}^n-\log P\left(y_{i} \mid \boldsymbol{Y}_{<i}\right)\right)
\end{equation}

To enforce left-to-right information flow, we apply an upper triangular attention mask that prevent attention from accessing future positions. As shown in the middle of Fig~\ref{fig:MTLM_model}, the shaded entries are set to $-\infty$. The final optimization goal is to minimize the sum of the cross-entropy between $h=\{h_1,...,h_5\}$ and the corresponding target sequence $y=\{y_2,...,y_6\}$, and $y_6$ is end token $eos$.

Note that since our model adopts an encoder-only architecture, once the hypothesis path is complete, the token probabilities for all positions can be computed in parallel within a single forward pass. This significantly reduces training costs and provides a substantial speed advantage for decoding strategies, especially for unidirectional n-best rescoring.

\subsection{Bidirectional Masked Language Modeling (BMLM)}

To enable the language model to better leverage right-context information and accelerate its learning of language patterns, we introduce the BMLM sub-task. The optimization objective is defined in Eq~\ref{eq7}, where $\boldsymbol{Y_{m}}$ denotes the set of masked tokens to be predicted, and $\boldsymbol{Y_{\backslash m}}$ represents the set of unmasked tokens in the sequence. As illustrated in the right part of Fig~\ref{fig:MTLM_model}, suppose $\boldsymbol{Y_{m}} = {y_2, y_5}$, Then, the model computes the representations ${h_1, h_4}$ for the masked positions, and the loss is defined as the sum of cross-entropy between these outputs and their corresponding ground truth tokens in $\boldsymbol{Y_{m}}$.

\begin{equation}
\label{eq7}
\mathcal{L}_{\mathrm{BMLM}}\left(\boldsymbol{Y},\boldsymbol{\theta}\right)=\mathbb{E}\left(\sum_{y_i \in \boldsymbol{Y}_{m}}-\log p\left(y_{i} \mid \boldsymbol{Y}_{\backslash m}\right)\right)
\end{equation}

In the BMLM task, the flow of information within the model is bidirectional: each token position can attend to all other positions except those corresponding to masked tokens. The attention mask is shown in the middle of Fig~\ref{fig:MTLM_model}. This allows the model to condition on richer contextual information when making predictions. Moreover, this bidirectional nature makes the model inherently compatible with bidirectional rescoring in decoding scenarios.

\subsection{Unidirectional Masked Language Modeling (UMLM)}

We design the UMLM auxiliary sub-task based on two key motivations. First, we aim to create an intermediate task that bridges the gap between ULM and BMLM, thereby reducing discrepancies among tasks in multi-task learning. This design facilitates knowledge transfer and sharing among different sub-tasks within the model. Second, when using a ULM-trained language model to assist ASR decoding, the prediction of the current token is heavily influenced by the accuracy of preceding predictions. This dependency reduce the robustness of the ASR system, especially when prior information is noisy or erroneous. To address this, we propose UMLM as an auxiliary task to simulate scenarios where the prior context is imperfect or partially missing.

The optimization objective of UMLM is defined in Eq~\ref{eq6}. Here $\boldsymbol{Y}_{r}$ denotes a set of randomly selected target tokens to be predicted, and $\boldsymbol{Y}_{m}$ is another randomly selected masked token. $\boldsymbol{Y}_{\backslash m}$ is still the unmasked token set. We ensure that for every target token in $\boldsymbol{Y}_{r}$, there is at least one token in its proceeding context selected to be included in $\boldsymbol{Y}_{m}$, such that the context is deliberately corrupted. The size of $\boldsymbol{Y}_{r}$ and  $\boldsymbol{Y}_{m}$ are equal. Fig~\ref{fig:MTLM_model} right part illustrates an example where $\boldsymbol{Y}_{r} = \{y_4, y_6\}$ and $\boldsymbol{Y}_{m} = \{y_2, y_4\}$. When predicting $y_4$, its left context $y_2$ is masked. When predicting $y_6$, ${y_2, y_4}$ are masked, further degrading the context quality. The final loss is computed as the sum of cross-entropy between the model outputs at masked positions ${h3, h5}$ and the corresponding ground truth tokens in $\boldsymbol{Y}_{r}$.

\begin{equation}
\label{eq6}
\mathcal{L}_{\mathrm{UMLM}}\left(\boldsymbol{Y},\boldsymbol{\theta}\right)=\mathbb{E}\left(\sum_{y_i \in \boldsymbol{Y}_{r}}-\log p\left(y_{i} \mid \boldsymbol{Y}_{<i} \cap \boldsymbol{Y}_{\backslash m} \right)\right)
\end{equation}

\subsection{MTLM Training Process}

The overall MTLM training consists of ULM, UMLM, and BMLM subtasks. The final loss function is defined in Eq~\ref{total_loss}. During training, each input sample inferences three times - once for each subtask - using the corresponding attention mask matrix. All three subtasks jointly contribute to the gradient updates. Specifically, the ULM subtask computes the cross-entropy loss between the model output and each token in the sequence. In contrast, UMLM and BMLM compute the cross-entropy loss only for masked tokens, which are sampled at a 30\% masking rate. 

\begin{equation}
\label{total_loss}
\min_{\theta} (\mathcal{L}_{\mathrm{ULM}}(\boldsymbol{Y},\theta)+\mathcal{L}_{\mathrm{UMLM}}(\boldsymbol{Y},\theta)+\mathcal{L}_{\mathrm{BMLM}}(\boldsymbol{Y},\theta))
\end{equation}

\subsection{Decoding}

The MTLM model can be applied to two major ASR decoding scenarios: shallow fusion and n-best rescoring \cite{ref9,ref11,ref_17_chorowski2016towards}. The rescoring stage can further be categorized into unidirectional and bidirectional modes. In this section, we describe how MTLM is utilized in each scenario.

\noindent {\bf{Shallow fusion}} integrates the language model (LM) scores with the acoustic model (AM) scores in real time during decoding. Specifically, during beam search, the LM scores are combined with AM scores in a weighted fashion to refine the scoring of each hypothesis (as shown in Eq~\ref{lm_integration_1}). Here $X$ is the speech input to the AM model, and $\lambda$ is the weight to control LM's impact to the decoding process. We adopt the one-pass decoding algorithm from \cite{ref25}, enabling joint scoring with the MTLM model and CTC+S2S AM. To reduce computational cost during inference, we first estimate the best transcription length using CTC greedy search \cite{ref25}. Simultaneously, we prune both completed and incomplete hypotheses to mitigate the bias towards shorter sequences advantages, which often leads to transcription errors.

\begin{equation}
\label{lm_integration_1}
\mathop{\arg \max}\limits_{y_i} (\log P_{AM}(y_i \mid X,\boldsymbol{Y}_{<i})+\lambda \log P_{LM}(y_i \mid \boldsymbol{Y}_{<i}))
\end{equation}

\noindent {\bf{N-best rescoring}} is performed after an initial decoding pass, where the top-n hypotheses from the ASR output are re-evaluated using the LM. MTLM model supports both unidirectional and bidirectional rescoring modes, corresponding to Eq~\ref{lm_integration_2} and Eq~\ref{lm_integration_3}, respectively. Here $\boldsymbol{Y}_{\backslash i}$ denotes the the rest tokens of the sentence except $y_i$. Since MTLM adopts an encoder-based architecture, all candidate hypotheses can be scored in a single forward pass, offering significant efficiency advantages over autoregressive models.

\begin{equation}
\label{lm_integration_2}
\operatorname{Score}_{\mathrm{LM}}(\boldsymbol{Y})= \sum_{i=1}^{\boldsymbol{n}} \log P_{\mathrm{LM}}\left({y}_{i} \mid \boldsymbol{Y_{<i}} ; \boldsymbol{\theta}\right)
\end{equation}

\begin{equation}
\label{lm_integration_3}
\operatorname{Score}_{\mathrm{LM}}(\boldsymbol{Y})= \sum_{i=i}^{\boldsymbol{n}} \log P_{\mathrm{LM}}\left({y}_{i} \mid \boldsymbol{Y}_{\backslash i} ; \boldsymbol{\theta}\right)
\end{equation}

\section{Experiment Details}

We evaluate our method on the LibriSpeech corpus \cite{ref26}, which contains 1000 hours of English speech along with corresponding transcriptions. The training set is partitioned into three subsets containing 100, 360, and 500 hours of audio, while the development and test sets are each categorized into clean and other conditions. The AM is trained using the transcribed speech data. For training the LM, we utilize both the transcriptions associated with the audio and the additional 800-million-word text-only corpus provided by LibriSpeech.

In our experiments, the AM follows the hybrid CTC+S2S architecture proposed in \cite{ref888}, where the CTC and sequence-to-sequence (S2S) components share a common encoder, while the S2S module has a dedicated decoder. The encoder comprises 12 Transformer blocks \cite{ref6}, each containing an 8-head self-attention layer and a 2048-dimensional feed-forward sub-layer. The decoder consists of 6 Transformer decoder layers. Input features are extracted using a 100-dimensional filter bank augmented with 3-dimensional pitch features. For subword modeling, we adopt byte-pair encoding (BPE) \cite{ref45} to construct a vocabulary of 7002 subword units, including special tokens such as $<sos>$, $<eos>$, and the CTC blank label.

Our MTLM model is built with six Transformer blocks, each equipped with 12 self-attention heads. The hidden size of the self-attention layers is 768, and the feed-forward network dimension is 3072. The MTLM shares the same 7002-sized vocabulary as the AM. The model is trained from scratch using the Adam optimizer with $\beta_1 = 0.9$ and $\beta_2 = 0.999$. We apply a learning rate schedule with 5000 warm-up steps, a peak learning rate of $2 \times 10^{-4}$, and a minimum learning rate decayed to $1 \times 10^{-6}$.

\section{Results and Discussion}
\begin{table}[ht]
  	\caption{The WER (\%) performance of UNILM and MTLM models using various decoding methods (lower is better). $\dagger$ means that the model uses Eq~\ref{lm_integration_2} for unidirectional $n$-best rescoring, $\blacklozenge$ means that the model uses Eq~\ref{lm_integration_3} for bidirectional $n$-best rescoring, and $\clubsuit$ means that the model uses the shallow fusion decoding. (Beam Size=3)
  	}
  \label{table1}
  \centering
  \begin{tabular}{ lccccc }
    \toprule
    \multirow{2}{*}{Model}  & \multicolumn{2}{c}{dev} & \multicolumn{2}{c}{ test } \\
        \cline { 2 - 5 }
  & clean$(\downarrow)$  & other$(\downarrow)$  & clean$(\downarrow)$ &  other$(\downarrow)$ \\
    \midrule
AM & 3.18 & 8.94 & 3.4 & 9.04 \\
    \midrule
+$\mathbf{UNILM}^{\dagger}$ &  3.02 & 8.69 & 3.18 & 8.78\\
+$\mathbf{UNILM}^{\clubsuit}$     & 2.57 & 6.99 & 2.81 & 7.36\\
    \midrule
+$\mathbf{MTLM}^{\dagger}$ & 3.01  & 8.67 & 3.14 & 8.75 \\
+$\mathbf{MTLM}^{\blacklozenge}$   & 3.03 & 8.67 & 3.14 & 8.76\\
+$\mathbf{MTLM}^{\clubsuit}$    & \textbf{2.46} & \textbf{6.75} & \textbf{2.63} & \textbf{7.08}\\
    \bottomrule
  \end{tabular}
\end{table}

In this section, we compare the word error rate (WER) performance of the ASR system when integrating the acoustic model (AM) with either a traditional unidirectional language model (UNILM) or our proposed MTLM, using both $n$-best rescoring and shallow fusion strategies. The UNILM is trained just using the ULM objective described in Section~\ref{sec:ULM}, which is commonly adopted in ASR systems. Table~\ref{table1} shows the WER results. The first row reports the baseline WER of the AM without any language model. The subsequent sections present the WERs after incorporating UNILM and MTLM under different decoding strategies. 

As shown in the table, integrating either UNILM or MTLM consistently improves performance over using the AM alone. Notably, MTLM model achieves superior performance across all settings. Specifically, MTLM achieves a WER of 2.63\% on test-clean and 7.08\% on test-clean and test-other, outperforming the UNILM. These results validate the effectiveness of introducing BMLM techniques into ASR language modeling.

Furthermore, we observe a consistent performance gap between $n$-best rescoring and shallow fusion. The reasons behind this are as follows: In $n$-best rescoring, the LM is applied after the AM has already generated a full set of hypotheses, limiting its influence to re-ranking only \cite{ref20,ref47}. In contrast, shallow fusion allows the LM to participate in the scoring process during beam search decoding, enabling joint hypothesis generation by the AM and LM. This tighter integration leads to improved WER performance, further demonstrating the advantage of using MTLM in a real-time ASR system \cite{ref_17_chorowski2016towards,ref18}.

\subsection{Error Type Analysis}

\begin{table}[ht]
\label{wer}
\vskip -13pt
  \caption{The number of error types for AM, UNILM and MTLM.}
  \label{table2}
  \centering
  \begin{tabular}{ ccccccc }
    \toprule
 Length & Model & Deletion$(\downarrow)$ & Insertion$(\downarrow)$ & Substitution$(\downarrow)$ & Overall$(\downarrow)$\\
    \midrule
     \multirow{5}{*}{L} & $\mathbf{AM}$ & 374 & 284 & 2650 & 3308\\
 & +$\mathbf{UNILM}^{\dagger}$ & 320 & 274 & 2541 & 3135  \\
 & +$\mathbf{MTLM}^{\dagger}$ & \textbf{319} & 273 & 2531 & 3123 \\
  & +$\mathbf{MTLM}^{\blacklozenge}$ & 320 & 272 & 2533 & 3125\\
 & +$\mathbf{UNILM}^{\clubsuit}$ & 370 & 205 & 2079 & 2654\\
 & +$\mathbf{MTLM}^{\clubsuit}$ & 367 & \textbf{199} & \textbf{1932} & \textbf{2498} \\
    \midrule 
    
     \multirow{5}{*}{M} & $\mathbf{AM}$ & 207 & 165 & 1620 & 1992\\
 & +$\mathbf{UNILM}^{\dagger}$ & 204 & 166 & 1565 & 1935 \\
 & +$\mathbf{MTLM}^{\dagger}$ & 203 & 164 & 1558 & 1925\\
   & +$\mathbf{MTLM}^{\blacklozenge}$ & 204 & 166 & 1559 & 1929 \\
 & +$\mathbf{UNILM}^{\clubsuit}$ & 203 & 116 & 1297 & 1616 \\
 & +$\mathbf{MTLM}^{\clubsuit}$ & \textbf{190} & \textbf{110} & \textbf{1260} & \textbf{1560}\\
    \midrule 
    
     \multirow{5}{*}{S} & $\mathbf{AM}$ & 150 & 104 & 975 & 1229 \\
 & +$\mathbf{UNILM}^{\dagger}$ & 154 & 105 & 943 & 1202 \\
 & +$\mathbf{MTLM}^{\dagger}$ & 153 & 103 & 936 & 1192 \\
   & +$\mathbf{MTLM}^{\blacklozenge}$ & 149 & 102 & 942 & 1193\\
 & +$\mathbf{UNILM}^{\clubsuit}$ & \textbf{135} & 89 & 855 & 1079 \\
 & +$\mathbf{MTLM}^{\clubsuit}$ & 143 & \textbf{78} & \textbf{826} & \textbf{1047} \\
  \bottomrule
  \end{tabular}
\vskip -15pt
\end{table}

In this section, we analyze transcription errors categorized into four types: deletion, insertion, substitution (replacement), and overall error. We combine the test-clean and test-other sets into a single dataset and categorize each utterance based on the number of words in its reference transcription. Specifically, we define three groups: short (fewer than 10 words), medium (10–20 words), and long (more than 20 words). Table~\ref{table2} summarizes the error counts for each model and decoding strategy. The AM row serves as the baseline.

As shown in Table 2, MTLM model outperforms UNILM in nearly all settings across all error types. The only exception occurs in the short category when using shallow fusion decoding, where MTLM produces a slightly higher total error count (143 errors) compared to UNILM (135 errors). Despite this, MTLM still demonstrates consistent improvements in the medium and long categories and achieves a notable reduction in the total number of overall errors across the dataset. These results confirm that MTLM is more robust than UNILM, particularly for longer utterances, where richer context allows its bidirectional and multi-task training advantages to be better leveraged. The findings also highlight MTLM’s superior generalization ability in real-world ASR scenarios involving diverse utterance lengths.

\subsection{MTLM Model's Influence in the Token Selection of Shallow Fusion}
During decoding, there is a $GuideScore$ algorithm which is responsible for selecting the next candidate tokens in each hypothesis path \cite{ref25}. This algorithm can operate either by using the joint scores from the AM S2S component and the LM, or by relying solely on the scores from the S2S component. In the experiments reported in Tables~\ref{table1} and \ref{table2}, the $GuideScore$ was configured to use only the AM S2S scores to guide decoding. To further investigate the impact of incorporating LM scores into this token selection process, we conducted ablation experiments to reveal the impact of MTLM model on $GuideScore$ algorithm. The WER results are presented in Table~\ref{table3}.

\begin{table}[ht]
  \caption{WER Results on introducing MTLM into $Guidescore$. (Beam Size=3)}
  \label{table3}
  \centering
  \begin{tabular}{ cccccc }
    \toprule
    \multirow{2}{*}{GuideScore} & \multirow{2}{*}{Decode Setting} & \multicolumn{2}{c}{ dev } & \multicolumn{2}{c}{ test } \\
        \cline { 3 - 6 }
 & &  clean$(\downarrow)$  &  other$(\downarrow)$  &  clean$(\downarrow)$  &  other$(\downarrow)$ \\
    \midrule
    \multirow{2}{*}{ S2S } & $\mathbf{UNILM}^{\clubsuit}$ &  2.57 & 6.99 & 2.81 & 7.36\\
    & $\mathbf{MTLM}^{\clubsuit}$ & \textbf{2.46} &  \textbf{6.75} & \textbf{2.63} & \textbf{7.08}\\
    \midrule
    \multirow{2}{*}{ MTLM+S2S } & $\mathbf{UNILM}^{\clubsuit}$ & 2.58  & 6.98 & 2.82 & 7.3 \\
    & $\mathbf{MTLM}^{\clubsuit}$ & \textbf{2.46} & 6.77 & 2.64 & 7.1\\
    \bottomrule
  \end{tabular}
\end{table}

As shown, when both the MTLM and S2S model contribute to the $GuideScore$, the WER of both the UNILM and MTLM models on the test-clean and test-other sets becomes slightly worse compared to when only the S2S scores are used. This outcome highlights a key insight: token selection during decoding should primarily reflect acoustic evidence from the speech signal, while LMs - being trained only on text - do not account for this situation. As such, excessive reliance on LM scores in this stage may introduce noise and hinder accurate hypothesis generation.

\subsection{The Effect of Different Beam Sizes}

\begin{figure}[htp]
  \centering
  \includegraphics[width=\linewidth]{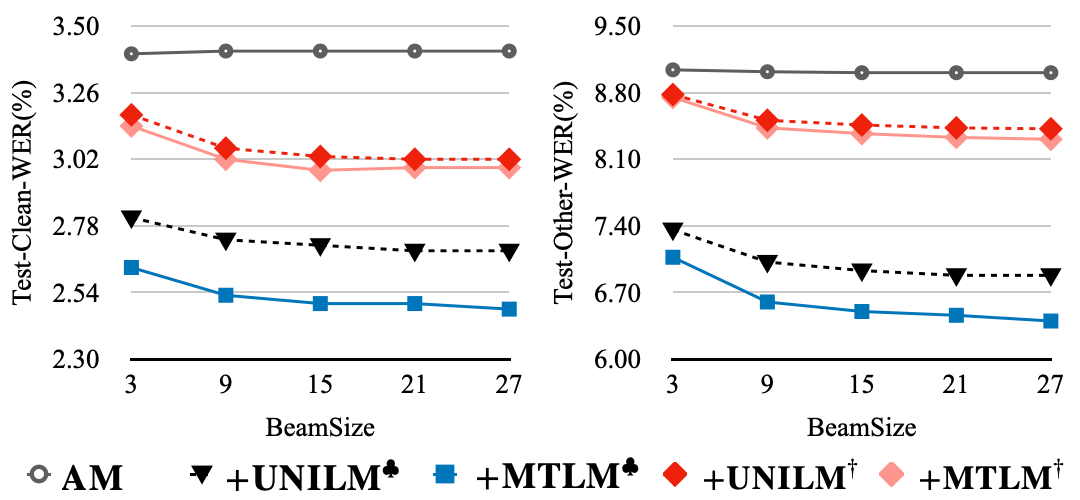}
  \caption{The WER performance of the AM, UNILM and MTLM on the \textit{test-clean} and \textit{test-other} datasets with different beam sizes.}
  \label{fig:speech_production}
\end{figure}

Figure 2 presents a comparison of word error rate (WER) among the acoustic model (AM), the UNILM, and the proposed MTLM model across varying beam sizes. The results demonstrate that, regardless of the decoding method employed, the MTLM consistently outperforms both the UNILM and the standalone AM. These results underscore the robustness of the MTLM under different decoding configurations and further highlight its effectiveness when integrated with beam search strategies in ASR systems.

Besides, increasing the beam size leads to a general reduction in WER for all models and decoding strategies. This is because larger beam sizes allow the decoder to explore a broader set of hypotheses, increasing the likelihood of selecting the correct transcription. Notably, the shallow fusion method exhibits the most substantial performance gains with increasing beam size, indicating that joint scoring from the AM and LM benefits more from expanded hypothesis space.

\section{Conclusion}

In this paper, we propose the MTLM training paradigm, which enhances the performance of language models in automatic speech recognition by incorporating bidirectional context during training. The resulting MTLM model is versatile and can be effectively applied to both shallow fusion and $n$-best rescoring decoding scenarios.

Experimental results on the LibriSpeech dataset show that MTLM consistently outperforms traditional unidirectional LMs (UNILM) across various decoding strategies. In addition, our error-type analysis demonstrates that MTLM achieves more significant reductions in transcription errors, particularly situation when richer context is required (for example, long utterance transcript). We further investigate the use of MTLM in alternative decoding settings, such as GuideScore, and examine its behavior under different beam sizes. These ablation studies underscore the robustness and effectiveness of MTLM in diverse ASR decoding contexts.

\bibliographystyle{splncs03_unsrt}
\bibliography{mybibliography}
\end{document}